# Spike and Tyke, the Quantized Neuron Model

M. A. El-Dosuky[1], M. Z. Rashad[1], T. T. Hamza[1], and A.H. EL-Bassiouny[2]

[1] Department of Computer Sciences, Faculty of Computers and Information sciences, Mansoura University, Egypt

mouh_sal_010@mans.edu.eg

magdi_12003@yahoo.com

Taher_Hamza@yahoo.com

[2] Department of Mathematics, Faculty of Sciences, Mansoura University, Egypt

el_bassiouny@mans.edu.eg

## Abstract

Modeling spike firing assumes that spiking statistics is Poisson, but real data violates this assumption. To capture non-Poissonian features, in order to fix the inevitable inherent irregularity, researchers rescale the time axis with tedious computational overhead instead of searching for another distribution! Spikes or action potentials are precisely-timed changes in the ionic transport through synapses adjusting the synaptic weight, successfully modeled and developed as a memristor. Memristance value is multiples of initial resistance. This reminds us with the foundations of quantum mechanics. We try to quantize potential and resistance, as done with energy.

After reviewing Planck curve for *blackbody radiation*, we propose the quantization equations. We introduce and prove a theorem that quantizes the resistance. Then we define the tyke showing its basic characteristics. Finally we give the basic transformations to model spiking and link an energy quantum to a tyke. Investigation shows how this perfectly models the neuron spiking., with over 97% match.

## 1. Introduction

For a long time, Hebb rule provides the basic conceptualization and learning algorithm for altering connection weights in neural network models [1]. Hebbian learning is only sensitive to the spatial correlations between pre- and postsynaptic neurons ([2], [3]). Spike-timing dependent plasticity (STDP) is temporally asymmetric form of Hebbian learning ([4], [5], [6]). STDP allows the demonstration of associative memory [9] and the construction of powerful brain-like circuitry ([7], [8]). There are many models for neuron spiking ([9], [10], [11], [12], [13]). Spikes or action potentials are precisely-timed changes in the ionic transport through synapses adjusting the synaptic weight [14]. A biological synapse successfully modeled and developed as a memristor in HP labs [15].

The Lagrange formalism of memory circuit elements shows how to approach the quantization of a general memory element circuit [16]. Memristors can be used to generate higher harmonics, as an example of the advantage of memristor circuits over circuits without memory[17].

The memristor is a state-based device, i.e., a pulse response depends on its starting point. Memristance at time t, denoted as $M_T$, is proved to be a function of flux $\phi$ [18],

$$M_T(t) = R_0 \sqrt{1 - \frac{2\eta \Delta R \phi(t)}{Q_0 R_0^2}} \qquad (1)$$

where, $R_0$ is the initial resistance, $\eta$ is the bias sign, $\Delta R$ is the difference between the maximum and the minimum resistance. $Q_0$ is the charge required to pass through the memristor.

According to the STDP rule, synaptic weight $w_{ij}$ between a pre-synaptic neuron j and a neuron i after a time step $\Delta t$ is, ([19],[20]):

$$w_{ij}(t + \Delta t) = w_{ij}(t) + \Delta w_{ij}(t) \qquad (2),$$

where

$$\Delta w_{ij} = \mu \, \text{sgn}(\hat{t}_i - \hat{t}_j) \exp(-|\hat{t}_i - \hat{t}_j|/\tau_d) \qquad (3)$$

where $\hat{t}_j$ and $\hat{t}_i$ are the relative timing of the pre- and post-synaptic spikes respectively, sgn(·) is the signum function, |·| is the absolute value, and $\tau_d$ is a reference time delay.

The problem statement of this research can be summarized as follows. Spike train generation assumes that spiking statistics is Poisson, but real data violates this assumption [21]. To capture non-Poissonian features, in order to fix the inevitable inherent irregularity, researchers rescale the time axis with tedious computational overhead ([22], [23]) instead of searching for another distribution!

We claim that we get the best model of spike train modeling and generation based on the following notice. From equation (1) we notice that memristance value is multiples of initial resistance. This reminds us with the foundations of quantum mechanics, when Planck found that atoms in a heated solid had energies that were multiples of a fixed amount, called a quantum [24]. Bohr explained the line spectrum of hydrogen, as in the energy diagram [24], shown in figure 1.

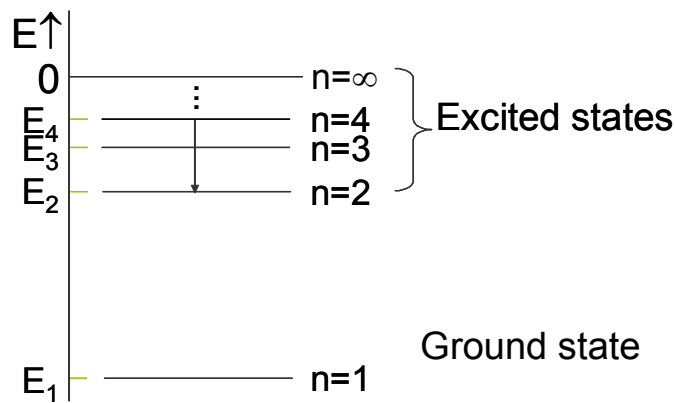

**Figure 1 Energy Diagram**

We try to quantize resistance, as done with energy. First we review Planck curve for *blackbody radiation*. Then we propose the quantization equations. Investigation shows how this perfectly models the neuron spiking, with the introduction of a new term called Tyke.

## 2. Planck curve for blackbody radiation

Planck investigated black-body radiation that hot objects glow. He derived an expression that perfectly describes the relation between the wavelength of radiation and the emitted power [25]:

$$P_\lambda = \frac{2\pi h c^2}{\lambda^5 (e^{hc/\lambda kT} - 1)} \qquad (4)$$

where $P_\lambda$ is power per m² area per m wavelength, h is Planck's constant (6.626 x $10^{-34}$ J.s), c is the speed of Light (3 x $10^8$ m/s), $\lambda$ is the wavelength (m), k is Boltzmann Constant (1.38 x $10^{-23}$ J/K), T is temperature (K). The curve is shown in Figure 2.

Planck assumed that atoms in the black body act as electromagnetic oscillators that emit radiation with a frequency $\nu$. These oscillators can only have energies given by [25]

$$E = nh\nu, \text{ n = 1, 2, 3…} \qquad (5)$$

Initially, at n =1, the smallest possible energy (h$\nu$) is the quantum of energy.

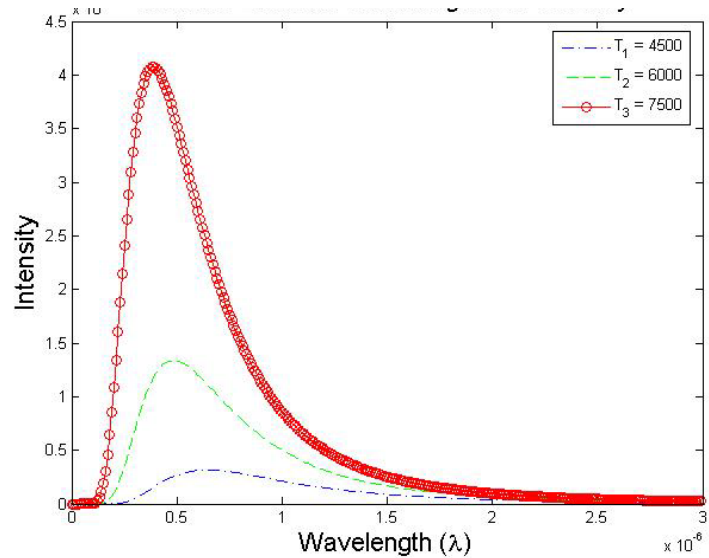
**Figure 2, Relation between Wavelength and Intensity**

The code listing that regenerates this curve is given below.
**Code listing 1: Plank Curve**
```
h = 6.6261*10^-34; % Planck's constant J. s
c = 2.9979*10^8; % speed of light m/s
k = 1.3807*10^-23; % Boltzmann's constant J/K
lambda = 1e-9:10e-9:3000e-9;
A1=(h.*c)./(k.* 4500.*lambda);
A2=(h.*c)./(k.* 6000.*lambda);
A3=(h.*c)./(k.* 7500.*lambda);
quantum =(8.*pi.*h.*c)./lambda.^5;
curve1= quantum.*(1./(exp(A1)-1));
curve2= quantum.*(1./(exp(A2)-1));
curve3= quantum.*(1./(exp(A3)-1));
plot(lambda, curve1,'-.b',lambda, curve2, 'g--',lambda, curve3, '-ro')
xlabel('Wavelength (\lambda)','FontSize',16)
ylabel('Intensity','FontSize',16)
title('\it{Relation Between Wavelength and Intensity}','FontSize',16)
hleg1 = legend('T_1 = 4500','T_2 = 6000', 'T_3 = 7500');
```

## 3. Spike and Tyke Model

Let us first introduce and prove the following theorem that quantizes the resistance. Then we define the tyke showing its basic characteristics. Finally we give the basic transformations to model spiking and link an energy quantum to a tyke.

**Theorem 1**: At any time, the resistance is $R = \dfrac{nh}{Q^2}$ where **n** = 1, 2, 3, …., an integer.

**Proof:** The power is the rate of generating or consuming energy.

$$P = \frac{E}{t} \qquad (6)$$

Power relates to current and voltage as it is equal to their multiplication at any time.
$$P = I \times V \qquad (7)$$
Using Ohm's Law: $V = I \times R$ and substituting in equation (7) to get:
$$P = I^2 \, R \qquad (8)$$
From equations (6) and (8) we get,
$$E = I^2 \, R \, t \qquad (9)$$
From equations (5) and (9) we get,
$$R = \frac{nh\upsilon}{I^2 t} \qquad (10)$$

Current is the rate of charge flow, and frequency is the reciprocal of the period denoted by T, so

$$R = \frac{nht^2}{Q^2 tT} = \frac{nht}{Q^2 T} = \frac{nh}{Q^2} \cdot \frac{t}{T} \quad (11)$$

where $\frac{t}{T}$ normalizes time with period time. Angular frequency is $2\pi\upsilon$, the angle $\theta$ is $2\pi\upsilon t = 2\pi \frac{t}{T}$, so $\frac{t}{T} = \frac{\theta}{2\pi}$. For $\theta = 0$, t = 0. For $\theta = 2\pi$, t = 1/$\upsilon$. If $0 \leq \theta \leq 2\pi$, then $0 \leq t \leq 1/\upsilon$. This can be seen as making the relative time or ranking the time over the phase. We omit this normalization or ranking part $\frac{t}{T}$, as the numbering variable *n* can undertake its role, so $R = \frac{nh}{Q^2}$ ∎

**Definition 1: Tyke**

Tyke is the potential corresponding to the smallest possible resistance.

Tyke can be calculated in different ways. Substituting n =1 in theorem 1, $R = \frac{nh}{Q^2}$, yields the smallest possible resistance ($\frac{h}{Q^2}$). Multiplying by the current to gain the tyke potential. We believe that tyke goes beyond measurement, but the measurable spike is actually the accumulation of tyke potentials over time.

**Transformation from Planck curve to Spike and Tyke Model**

Transforming the space of plank curve to the space of neuron potential is straight foreward, by defining two transformation functions.

$$f_x: \text{wavelength} \rightarrow \text{time}, \qquad f_x(x) = \frac{x}{c} \quad (12)$$

$$f_y: \text{Intensity} \rightarrow \text{potential}, \qquad f_y(y) = y\frac{A}{I} \quad (13)$$

Function $f_x$ is based on $\frac{1}{T} = \upsilon = \frac{c}{\lambda}$, so the time is gained by dividing the wavelength by c, the speed of light. Function $f_y$ is based on $\text{Intensity} = \frac{IV}{A}$, so the potential is gained by dividing the Intensity by *I* and multiplying by the area *A*. Successful generation of spike train is shown in figure 3. The code listing that regenerates this curve is given below.

**Code listing 2: Spiking potential**

```
h = 6.6261*10^-34;
c = 2.9979*10^8;
k = 1.3807*10^-23;
lambda_tyke = 1e-9:10e-9: 3000e-9;
lambda = 1e-9:10e-9: 8* 3000e-9;
time=  lambda ./ c;
qu=(8.*pi.*h.*c)./lambda_tyke.^5;
figure(1),clf
spike=zeros([1, 300]);
for ii= 4500: 500 :7500
   Ai=(h.*c)./(k.*ii.*lambda_tyke);
   sgn=qu.*(1./(exp(Ai)-1))
   spike = cat( 2, spike, sgn);
end
plot(t,spike, '-g', 'LineWidth',2)
hold on
xlabel('Time','FontSize',16)
ylabel('Potential','FontSize',16)
```

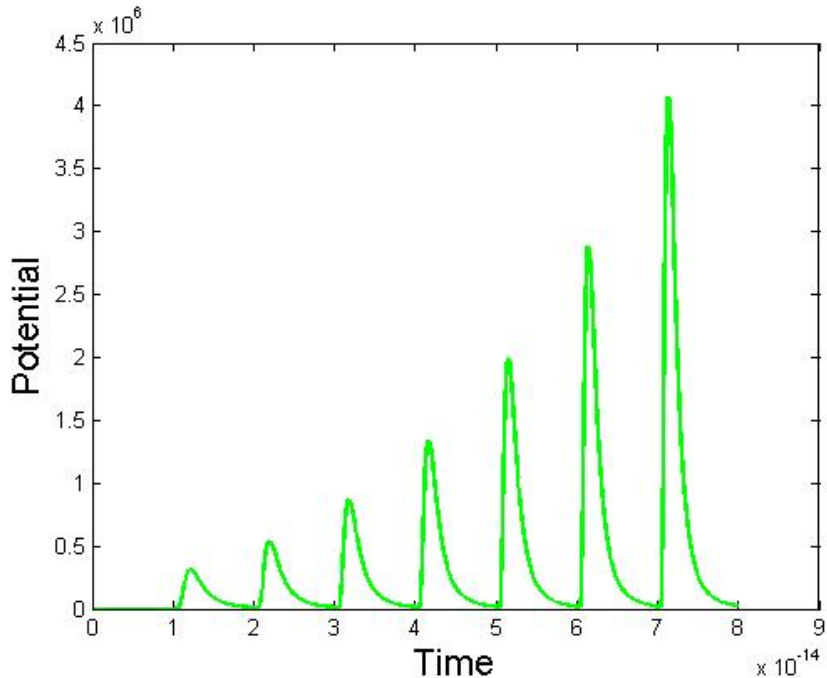

**Figure 3, generated spike train, showing relation between time and potential**

## 4. Investigation and Future work

To evaluate the proposed model, we run a simulation, with time between $t_0 = 3.3357 \times 10^{-18}$ sec to $t_{max} = 9.9770 \times 10^{-15}$ sec, and step size $\Delta t = 3.3357 \times 10^{-17}$ sec. That is about 300 points. We use a simple performance measure of counting the matched points. On average we have 280 matched points. This is over 97% match of the curve points, as shown in figure 4.

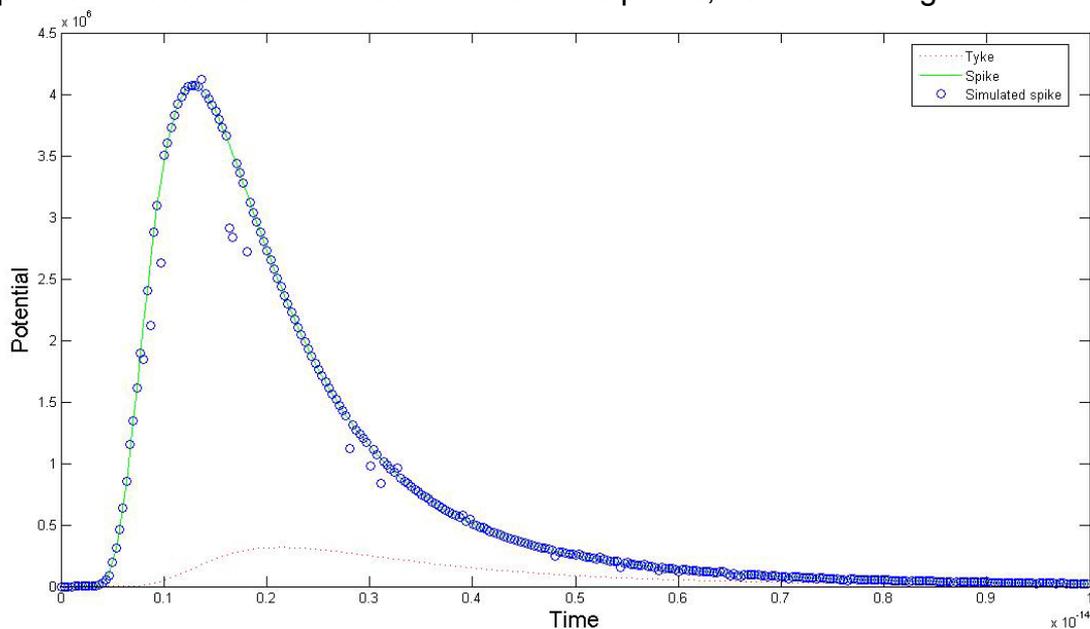

**Figure 4, evaluating the proposed model by counting matched points.**

Future work has many dimensions. One possible direction is answering the question: What is the relation between the class of problems solved by the current model and the class of problems solved by chaotic stochastic models? If the two classes are equivalent or there is a mapping between them, there would be no need to hybridize neural network with fuzzy sets, rough sets, or chaos. Another future work direction is investigating how to exploit other quantum physics equations, such as Wien's. Wien proposed an approximation for the spectrum of the object, which was correct at high frequencies (short wavelength) but not at low frequencies (long wavelength) [25]. Finally, a direction is to seek an optimal learning rule the spiking neuron model, to enhance operation of artificial neural networks.